\documentclass[twoside,11pt]{article}
\usepackage{jmlr2eedit}
\usepackage{amsmath}
\usepackage{amssymb}
\usepackage{graphicx}
\usepackage{times}
\usepackage{helvet}
\usepackage{courier}
\usepackage{epsfig}
\usepackage{subfigure}
\usepackage{makeidx}
\usepackage{multirow}
\usepackage{url}

\begin{document}

\title{Text Modeling using Unsupervised Topic Models and Concept Hierarchies}

\author{\name Chaitanya Chemudugunta \email chandra@ics.uci.edu \\
       \addr Department of Computer Science \\
University of California, Irvine \\
Irvine, CA, USA
        \AND
        \name Padhraic Smyth \email smyth@ics.uci.edu \\
       \addr Department of Computer Science\\
        University of California, Irvine \\
        Irvine, CA, USA
        \AND
       \name Mark Steyvers \email msteyver@uci.edu \\
       \addr Department of Cognitive Sciences\\
        University of California, Irvine\\
        Irvine, CA, USA}

\maketitle
\begin{abstract}
Statistical topic models provide a general data-driven
framework for automated discovery of high-level knowledge from large
collections of text documents. While topic models
can potentially discover a broad range of themes in a data set,
the interpretability of the learned topics is not always ideal. Human-defined
concepts, on the other hand, tend to be semantically richer
due to careful selection of words to define concepts
but they tend not to cover the themes in a data
set exhaustively. In this paper, we propose a
probabilistic framework to combine a hierarchy of
human-defined semantic concepts with statistical topic models
to seek the best of both worlds. Experimental results
using two different sources of concept hierarchies and
two collections of text documents indicate that this
combination leads to systematic improvements in the quality
of the associated language models as well as enabling new techniques
for inferring and visualizing the semantics of a document.
\end{abstract}

\section{Introduction}

There are a variety of popular and useful techniques for
automatically summarizing the thematic content of a set of
documents including document clustering \citep{nigam00text} and latent semantic
analysis   \citep{LandDum97}. A somewhat more recent
and general framework that has been developed in this
context is  latent Dirichlet analysis \citep{BleiNg03}, also
referred to as statistical topic modeling \citep{GriffStey04}.
 The basic concept underlying statistical
topic modeling is that each document is composed of a probability
distribution over topics, where each topic is represented as a
multinomial probability distribution over words.   The
document-topic and topic-word distributions are learned
automatically from the data in an unsupervised manner with no
human labeling or prior knowledge required. The underlying
statistical framework of topic modeling  enables a variety of
interesting extensions to be developed in a systematic manner,
such as author-topic models \citep{steyversat},  correlated
topics \citep{BleiLafferty05}, and hierarchical topic models
\citep{NCRPBlei07,Mimnohpam07}.

\begin{table}[htdp]
\begin{center} 
\begin{tabular*}{0.4\textwidth}{@{\extracolsep{\fill}} c||cc}
Word  & Topic A & Topic B\\
\hline
 database & 0.50  & 0.01 \\
 query & 0.30 & 0.01 \\
 algorithm & 0.18 & 0.08 \\
 semantic & 0.01 & 0.40 \\
 knowledge & 0.01 & 0.50 \\
\end{tabular*}
\end{center}
\caption{Toy example illustrating 2 topics each with 5 words.}
\label{tab:topics} 
\end{table}

As an illustrative  example, Table \ref{tab:topics} shows two
example topics defined over a toy vocabulary with 5 words
Individual documents could then be modeled as coming entirely from
topic A or from topic B, or more generally as a mixture (50-50,
70-30, 10-90, etc) from the two topics.

The topics learned by a topic model can be thought of as themes
that are discovered from a corpus of documents, where the
topic-word distributions ``focus" on the high probability words
that are relevant to a theme. An entirely different approach is to
manually define semantic concepts using human knowledge and
judgement. In the construction of ontologies and thesauri it is
typically the case that for each concept a relatively small set of
important words  associated with the concept are defined based on
prior knowledge. Concept names and sets of relations among
concepts (for ontologies) are also often provided.

\begin{table}[htdp]
\begin{center} 
\begin{tabular*}{0.4\textwidth}{@{\extracolsep{\fill}} c||cc}
\textsc{Family} Concept\hspace{1.5mm} & \multicolumn{2}{c}{\textsc{Family} Topic}\\
\hline
beget & family &(0.208) \\
birthright & child &(0.171) \\
brood & parent &(0.073) \\
brother & young &(0.040) \\
children & boy &(0.028) \\
distantly & mother &(0.027) \\
dynastic & father &(0.021) \\
elder &  school &(0.020) \\
\end{tabular*}
\end{center}
\caption{CALD \textsc{Family} concept and learned \textsc{Family} topic}
\label{tab:family} 
\end{table}

Concepts (as defined by humans) and  topics  (as learned from
data)  represent similar information but in different ways.  As an
example, the left column in Table~\ref{tab:family} lists some of
the 204 words that have been manually defined as part of  the
concept \textsc{Family} in the Cambridge Advanced Learners
Dictionary (more details on this set of concepts are provided
later in the paper). The right column shows the high probability
words for a learned topic, also about families. This topic was
learned automatically from a text corpus using a statistical topic
model. The numbers in parentheses are the probabilities that a
word will be generated conditioned on the learned topic---these
probabilities sum to 1 over the entire vocabulary of words,
specifying a multinomial distribution. The concept \textsc{Family}
in effect puts probability mass 1 on the set of 204 words within
the concept, and probability 0 on all other words. The topic
multinomial on the other hand could be viewed as a ``soft" version
of this idea, with non-zero probabilities for all words in the
vocabulary---but significantly skewed, with most of the
probability mass focused on a relatively small set of words.

Human-defined concepts are likely to be more interpretable than
topics and can be broader in coverage, e.g., by including words
such as {\it beget} and {\it brood} in the concept \textsc{Family}
in Table 1. Such relatively rare words will occur rarely (if at
all) in a particular corpus and are thus far less likely to be
learned by the topic model as being associated with the more
common family words.

Topics on the other hand have the advantage of being
tuned to the themes in the particular corpus they are trained on.
In addition, the probabilistic model that underlies the topic
model allows one to automatically tag each word in a document with
the topic most likely to have generated it. In contrast, there are
no general techniques that we are aware of that can automatically
tag words in a document with relevant concepts from an ontology or
thesaurus.

In this paper we propose a general framework for combining
data-driven topics and semantic concepts, with the goal of taking
advantage of the best features of both approaches.  Section \ref{sec:datasets} describes the two large ontologies and the text corpus
that we use as the basis for our experiments.  We begin
Section \ref{sec:ctm} by reviewing the basic principles of topic models and
then introduce the concept-topic model which combines   concepts
and  topics into a single probabilistic model. In Section \ref{sec:hctm} we
then extend the framework to the hierarchical concept-topic model
to take advantage of known hierarchical structure among concepts.
In Section \ref{sec:illustrations} we
discuss a number of examples that illustrate how the hierarchical
concept-topic model works, showing for example how an ontology can
be matched to a corpus and how
 documents can be tagged at the word-level with concepts from an
 ontology. Section \ref{sec:experiments} describes a series of experiments that
 evaluate the predictive performance of a number of different
 models, showing for example that prior knowledge of concept words and
 concept relations can lead to better topic-based language models.
 Sections \ref{sec:future} and \ref{sec:conclusions} conclude the paper with
 a brief discussion of future directions and final comments.

In terms of related work, our approach builds on the general
 topic modeling framework of
 \cite{BleiNg03} and
 \cite{GriffStey04} and  the hierarchical Pachinko
 models of   \cite{Mimnohpam07}. Almost all earlier work
 on topic modeling is purely data-driven in that no human
 knowledge is used in learning the topic models. One exception is the work by 
  \cite{IfrimTW-ICML2005} who apply the aspect 
 model \citep{Hofmann01} to model background knowledge in the 
 form of concepts to improve text classification.  Another exception
is the work of \cite{BoydBlei07} who
develop a topic modeling framework that
  combines human-derived linguistic knowledge with unsupervised
topic models for the purpose of word-sense disambiguation. Our
framework is somewhat more general than both of these approaches in that we not only improve the
quality of making predictions on text data by using prior human
concepts and concept-hierarchy, but also are able to make inferences in the reverse
direction about concept words and hierarchies given data.

There is also a significant amount of prior work on using data to
help with ontology construction and evaluation, e.g., learning
ontologies from text data (e.g., \cite{Maedche01}) or
methodologies for evaluating how well ontologies are matched to
specific text corpora ~\citep{BrewWilks04,AlanBrew06}. Our work is
broader in scope in that we propose general-purpose probabilistic
models that combine concepts and topics within a single framework,
allowing us to use the data to make inferences about how documents
and concepts are related (for example). It should be noted that in
the work described in this paper we do not explicitly investigate
techniques for modifying an ontology in a data-driven manner
(e.g., adding/deleting words from concepts or relationships among
concepts)---however, the framework we propose could certainly be
used as a basis for exploring such ideas.

\section{Text Data and Concept Sets } \label{sec:datasets}

The experiments in this paper are based on one large text corpus and two
different concept sets. For the text corpus, we used the Touchstone Applied
Science Associates (TASA) dataset \citep{LandDum97}. This corpus consists of
$D=37,651$ documents with passages excerpted from educational texts used in
curricula from the first year of school to the first year of college. The
documents are divided into 9 different educational genres. In this paper, we
focus on the documents classified as \textsc{Science} and \textsc{Social
Studies}, consisting of $D=5,356$ and $D=10,501$ documents and 1.7 Million and 3.4 Million
word tokens respectively.

For human-based concepts the first source we used was  a
thesaurus from the Cambridge Advanced Learner's Dictionary
(CALD; http://www.cambridge.org/elt/dictionaries/cald.htm). CALD consists of $C=2,183$
hierarchically organized semantic categories. In contrast to other
taxonomies such as WordNet \citep{fellbaum98}, CALD groups words
primarily according to semantic topics with the topics
hierarchically organized. The hierarchy starts with the concept
\textsc{Everything} which splits into 17 concepts at the second
level (e.g. \textsc{Science}, \textsc{Society},
\textsc{General/Abstract}, \textsc{Communication}, etc). The
hierarchy has up to 7 levels.  The concepts vary in the number of the words
with a median of 54 words and a maximum of 3074. Each word can be
a member of multiple concepts, especially if the word has multiple
senses.

The second source of concepts in our experiments was the
Open Directory Project (ODP), a human-edited hierarchical
directory of the web (available at http://www.dmoz.org). The ODP
database contains descriptions and URLs on a large number of
hierarchically organized topics. We extracted all the topics in
the \textsc{Science} subtree, which consists of $C=10,817$ concept nodes
after preprocessing. The top concept in this hierarchy starts with
\textsc{Science} and divides into topics such as
\textsc{Astronomy}, \textsc{Math}, \textsc{Physics}, etc. Each of
these topics divides again into more specific topics with a
maximum number of 11 levels. Each node in the hierarchy is
associated with a set of URLs related to the topic plus a set of
human-edited descriptions of the site content. To create a bag of
words representation for each node, we collected all the words in
the textual descriptions and also crawled the URLs associated with
the node (a total of 78K sites). This led to a vector of word
counts for each node.

For both the concept sets, we propagate the words upwards in the
concept tree so that an internal concept node is associated with
its own words and all the words associated with its children.  We
created a single $W = 21,072$ word vocabulary based on the 3-way
intersection between the vocabularies of TASA, CALD, and ODP. This
vocabulary covers approximately 70\% of all of the word tokens in the TASA
corpus and is the vocabulary that is used in all of the
experiments reported in this paper. We also generated the same set
of experimental results using the union of words in TASA,
CALD, and ODP, and found the same general behavior as with the
intersection vocabulary. We report the intersection results 
and omit the union results as they are essentially identical to
the intersection results. A useful feature of using the
intersection is that it  allows us to evaluate two different sets
of concepts (CALD and ODP) on a common data set (TASA) and
vocabulary.

\section{Combining Concepts and Topics } \label{sec:ctm}

In this section, we describe the concept-topic model and
detail its generative process and describe an
illustrative example.  We first begin with a
brief review of the topic model.

\subsection{Topic Model} \label{subsec:tm}

The topic model (or latent Dirichlet allocation model) is
a statistical learning technique for extracting a set of topics that
describe a collection of documents \citep{BleiNg03}.  A
topic $z$ is represented as a multinomial distribution over the $V$ unique
words in a corpus, $p(w|z) = [p(w_1|z), ..., p(w_V|z)]$ such
that $\sum_v p(w_v|z) = 1$.  Therefore, a topic can be viewed a $V$-sided die and
generating $n$ words from a topic is akin to throwing the topic-die $n$ times.  There
are a total of $T$ topics and a document $d$ is represented as
a multinomial distribution over those $T$ topics $p(z|d)$, $1 \le z \le T$ and $sum_z p(z|d) = 1$.  Generating a word
from a document $d$ involves first selecting a topic $z$ from the
document-topic distribution $p(z|d)$ and then
selecting a word from the topic distribution $p(w|z)$.  This process is repeated for each word in the
document.  The conditional probability of a word in a document is given by,

\begin{equation}
p(w|d) = \sum_{z} p(w|z) p(z|d) \label{eqn:mixture}
\end{equation}

Given the words in a corpus, the inference problem involves
estimating the word-topic distributions $p(w|z)$ and the
topic-document distributions $p(z|d)$ for the corpus.  For the
standard topic model, collapsed Gibbs sampling has been
successfully applied to do inference on large text collections in
an unsupervised fashion \citep{GriffStey04}.  Under this technique,
words are initially assigned randomly to topics and the algorithm
then iterates through each word in the corpus and samples a topic
assignment given the topic assignments of all other words in the
corpus.  This process is repeated until a steady state is reached
(e.g. the likelihood of the model on the corpus is not increasing
with subsequent iterations) and the topic assignments to words are
then used to estimate the word-topic $p(w|z)$ and topic-document
$p(z|d)$ distributions.  The topic model uses Dirichlet priors on
the multinomial distributions $p(w|z)$ and $p(z|d)$.  In this
paper, we use a fixed symmetric prior on $p(w|z)$ word-topic
distributions and optimize the asymmetric Dirichlet prior
parameters on $p(z|d)$ topic-document distributions using fixed
point update equations (as given in \cite{Minka00}). See Appendix
A for more details on inference.

\subsection{Concept-Topic Model} \label{subsec:ctm}

\begin{figure*} \centering
\includegraphics[keepaspectratio,width=5.7in]{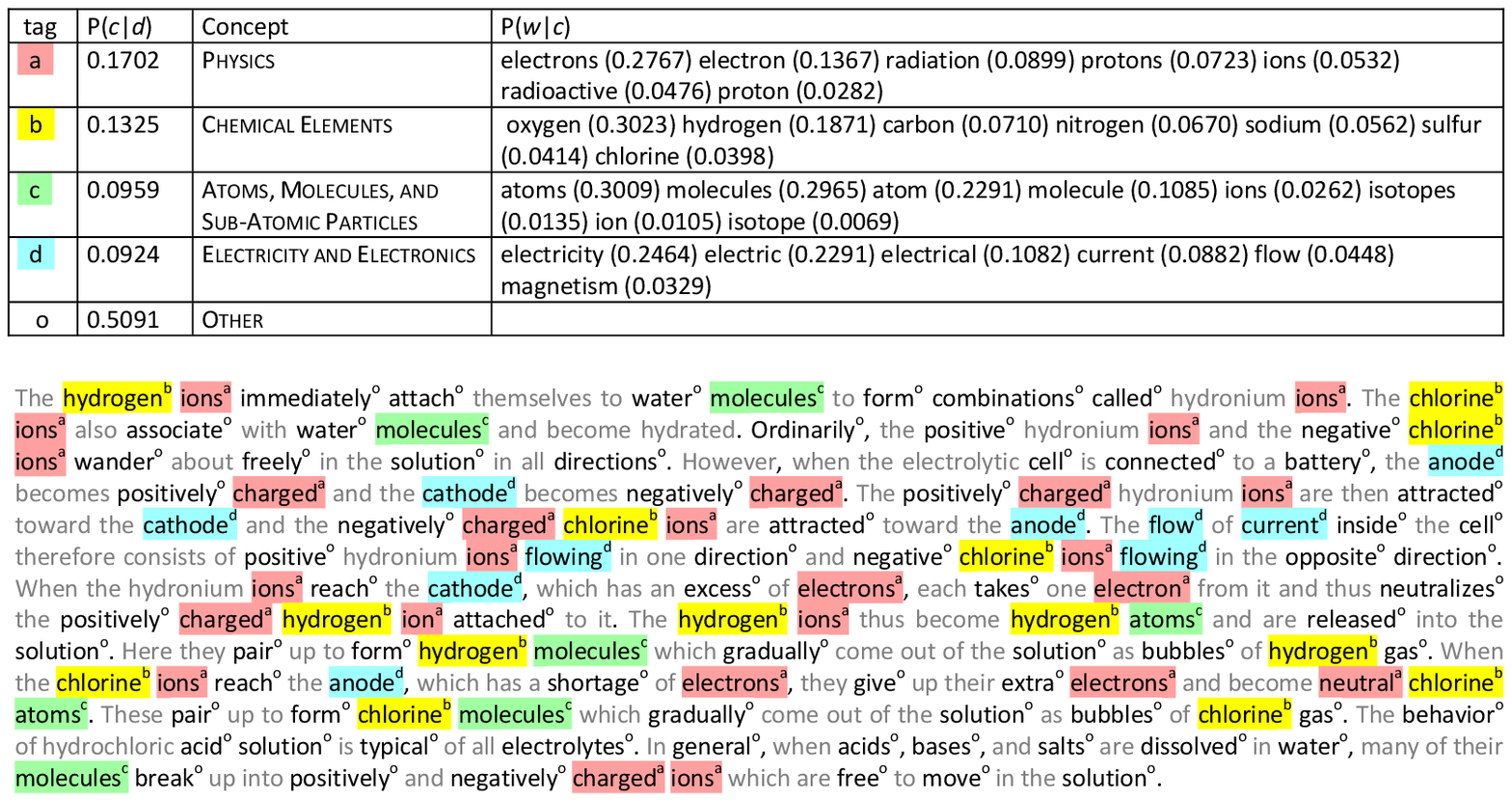}
\caption{Illustrative example of tagging a document excerpt using
the concept model (CM) with concepts from CALD.
\label{fig:caldtag}}
\end{figure*}

The concept-topic model is a simple extension to the topic model where we add
$C$ concepts to the $T$ topics of the topic model resulting in an effective set of
$T$ + $C$ ``topics" for each document.

Recall that a concept is represented as a
set of words.  The human-defined concepts only give us a membership function over
words---either a word is a member of the concept or it
is not.
One straightforward way to incorporate concepts
into the topic modeling framework is
to convert them to ``topics" by representing them as probability distributions
over their associated word sets.  In other words, a concept $c$ can be represented by a
multinomial distribution $p(w|c)$ such that $\sum_w p(w|c) = 1$ where $w \in c$ (therefore,
$p(w|c) = 0$ for $w \notin c$).  A document is now represented as a distribution over
topics and concepts, $p(z|d)$ where $1 \le z \le T+C$.  The conditional probability of
a word $w$ given a document $d$ is,

\begin{equation}
p(w|d) = {\sum_{t=1}}^{T} p(w|t) p(t|d) + {\sum_{c=1}}^{C} p(w|c) p( T + c |d) \label{eqn:ctmmixture}
\end{equation}

The generative process for a document collection with $D$
documents under the concept-topic model is as follows:

\begin{enumerate}
  \item For each topic $t \in \{1,...,T\}$, select a word distribution $\phi_t$ $\sim$ Dir($\beta_{\phi}$)
  \item For each concept $c \in \{1,...,C\}$, select a word distribution $\psi_c$ $\sim$ Dir($\beta_{\psi}$) \footnote{Note that $\psi_c$ is a constrained
  word distribution defined over only the words that are members of the human-defined concept $c$}
  \item For each document $d \in \{1,...,D\}$
  \begin{enumerate}
    \item Select a distribution over topics and concepts $\theta_d$ $\sim$ Dir($\alpha$)
    \item For each word $w$ of document $d$
    \begin{enumerate}
      \item Select a component $z$ $\sim$ Mult($\theta_d$)
      \item If $z \le T$ generate a word from topic $z$, $w$ $\sim$ Mult($\phi_{z}$); otherwise generate
    a word from concept $c = z-T$, $w$ $\sim$ Mult($\psi_{c}$)
    \end{enumerate}
  \end{enumerate}
\end{enumerate}

where $\phi_t$ represents the $p(w|t)$ word-topic
distribution for topic $t$, $\psi_c$ represents
the $p(w|c)$ word-concept distribution for concept $c$ and $\theta_d$ represents the
$p(z|d)$ distribution over topics and concepts for document $d$.  $\beta_{\phi}$,
$\beta_{\psi}$ and $\alpha$ are the parameters of the Dirichlet priors for $\phi$, $\psi$ and $\theta$
respectively.  Every element in the above process is unknown except for the
words in the corpus and the membership of words in the
human-defined concepts.  Thus, the inference problem involves estimating the
distributions $\phi$, $\psi$ and $\theta$ given the words in the corpus.  The standard collapsed
Gibbs sampling scheme previously used to do inference for the topic model can be modified
to do inference for the concept-topic model.  We also optimize the Dirichlet parameters
using the fixed point updates from \cite{Minka00} after each Gibbs sampling sweep through
the corpus.

The topic model can be viewed as a special case of the concept-topic model when there
are no concepts present, i.e. when $C = 0$.  The other extreme of this model
where $T = 0$, which we refer to as the concept model, is used for illustrative purposes.  In our experiments,
we refer to the topic model, concept model and the concept-topic model as TM, CM and CTM respectively.

We note that the concept-topic model is not the only way to
incorporate semantic concepts.  For example, we could use the
concept-word associations to build informative priors for the
topic model and then allow the inference algorithm to learn word
probabilities for all words (for each concept), given the prior
and the data.  We chose the current approach to exploit the
sparsity in the concept-word associations (topics are
distributions over all the words in the vocabulary but concepts
are restricted to just their associated words). This allows us to
easily do inference with tens of thousands of concepts on large
document collections.  A motivation for this approach is that
there might be topics present in a corpus (that can be learned)
that are not represented in the concept set. Similarly, there may
be concepts that are either missing from the text corpus or are
rare enough that they are not found in the data-driven topics of
the topic model.  This marriage of concepts and topics provides a
simple way to augment concepts with topics and has the flexibility
to mix and match topics and concepts to describe a document.

Figure \ref{fig:caldtag} illustrates concept assignments to individual
words in a TASA document
with CALD concepts, using the concept model (CM).  The four most
likely concepts are listed for this document.  For each concept,
the estimated probability distribution over words is shown next to
the concept. In the document, words assigned to the four most
likely concepts are tagged with letters a-d (and color coded if
viewing in color). The words assigned to any other concept are
tagged with ``o" and words outside the vocabulary are not tagged.
In the concept model, the distributions over concepts within a
document are highly skewed such that most probability goes to only
a small number of concepts. In the example document, the four most
likely concepts cover about 50\% of all words in the document.

The figure illustrates that the model correctly disambiguates
words that have several conceptual interpretations. For example,
the word \textit{charged}  has many different meanings and appears
in 20 CALD concepts. In the example document, this word is
assigned to the PHYSICS concept which is a reasonable
interpretation in this document context. Similarly, the ambiguous
words \textit{current} and \textit{flow} are correctly assigned to
the ELECTRICITY concept.

\section{Hierarchical Concept-Topic Model } \label{sec:hctm}

Concepts are often arranged in a tree-structured hierarchy.  While the concept-topic model provides a
simple way to combine concepts and topics, it does not take into account the
hierarchical structure of the concepts.  In this section, we describe an extension, the
hierarchical concept-topic model, that extends the concept-topic model to incorporate the
hierarchical structure of the concept set.

Similar to the concept-topic model described in the previous section, there are $T$ topics and
$C$ concepts in the hierarchical concept-topic model.  For each document $d$, we introduce a
``switch" distribution $p(x|d)$ which determines if a word should be generated via the topic route or the
concept route.  Every word token in the corpus is associated with a binary switch variable $x$.  If
$x$ = 0, the previously described standard topic mechanism of Section \ref{subsec:tm} is used to generate the word.  That is,
we first select a topic $t$ from a document-specific mixture of topics $p(t|d)$ and generate
a word from the word distribution associated with topic $t$.  If $x$ = 1, we generate the word
from one of the $C$ concepts in the concept tree.  To do that, we associate with
each concept node $c$ in the concept tree a document-specific multinomial distribution with
dimensionality equal to $N_c$ + 1, where $N_c$ is the number of children of the concept
node $c$.  This distribution allows us to traverse the concept tree and exit at any of the
$C$ nodes in the tree --- given that we are at a concept node $c$, there are $N_c$ child concepts to
choose from and an additional option to choose an ``exit" child to exit the concept tree
at concept node $c$.  We start our walk through the concept tree at the root
node and select a child node from one of its children.  We repeat this process until
we reach an exit node and the word is generated from the the parent of the exit node.  Note that
for a concept tree with $C$ nodes, there are exactly $C$ distinct ways to select a path and exit the
tree --- one for each concept.

In the hierarchical concept-topic model, a document is represented as a 
weighted combination of mixtures of $T$ topics and $C$ paths through the concept tree
and the conditional probability of a word $w$ given a document $d$ is given by,

\begin{eqnarray}
p(w|d) = P(x=0|d) \sum_{t} p(w|t) p(t|d) ~~~~~~~~~~~ \nonumber \\
~~~~~~~~~~~~~~~~~~~~~~~~~~~~ + P(x=1|d) \sum_{c} p(w|c) p(c|d) \label{eqn:hctmmixture}
\end{eqnarray}

where  $~~~~p(c|d) = p(exit|c) p(c|parent(c))...p(.|root)$
\\\\
The generative process for a document
collection with $D$ documents under the hierarchical concept-topic model is as follows: -

\begin{enumerate} 
  \item For each topic $t \in \{1,...,T\}$, select a word distribution $\phi_t$ $\sim$ Dir($\beta_{\phi}$)
  \item For each concept $c \in \{1,...,C\}$, select a word distribution $\psi_c$ $\sim$ Dir($\beta_{\psi}$) \footnote{Note that $\psi_c$ is a constrained
  word distribution defined over only the words that are members of the human-defined concept $c$}
  \item For each document $d \in \{1,...,D\}$
  \begin{enumerate} 
    \item Select a switch distribution $\xi_d$ $\sim$ Beta($\gamma$)
    \item Select a distribution over topics $\theta_d$ $\sim$ Dir($\alpha$)
    \item For each concept $c \in \{1,...,C\}$
    \begin{enumerate} 
      \item Select a distribution over children of $c$, $\zeta_{cd}$ $\sim$ Dir($\tau_c$)
    \end{enumerate} 
    \item For each word $w$ of document $d$
    \begin{enumerate} 
      \item Select a binary switch variable $x$ $\sim$ Bernoulli($\xi_d$)
      \item If $x$ = 0
      \begin{enumerate} 
        \item Select a topic $z$ $\sim$ Mult($\theta_d$)
    \item Generate a word from topic $z$, $w$ $\sim$ Mult($\phi_{z}$)
      \end{enumerate} 
      \item Otherwise, create a path starting at the root concept node, $\lambda_1$ = 1
      \begin{enumerate} 
    \item Repeat \\
      {\small Select a child of node $\lambda_j$, $\lambda_{j+1}$ $\sim$ Mult($\zeta_{{\lambda_j}{d}}$)} \\
    Until $\lambda_{j+1}$ is an exit node
    \item Generate a word from concept $c$ = $\lambda_{j}$, $w$ $\sim$ Mult($\psi_{c}$); set $z$ to $T + c$
      \end{enumerate} 
    \end{enumerate} 
  \end{enumerate} 
\end{enumerate} 

where $\phi_t$, $\psi_c$, $\beta_{\phi}$ and $\beta_{\psi}$ are
analogous to the corresponding symbols in the concept-topic model
described in the previous section.  $\xi_{d}$ represents the
$p(x|d)$ switch distribution for document $d$, $\theta_d$
represents the $p(t|d)$ distribution over topics for document $d$,
$\zeta_{cd}$ represents the multinomial distribution over children
of concept node $c$ for document $d$ and $\gamma$, $\alpha$,
$\tau_c$ are the parameters of the priors on $\xi_d$, $\theta_d$,
$\zeta_{cd}$ respectively.  As before, all elements above are
unknown except words and the word-concept memberships in the
generative process.  Details of the inference technique based on
collapsed Gibbs sampling \citep{GriffStey04} and fixed point update
equations to optimize the Dirichlet parameters \citep{Minka00} are
provided in  Appendix A.

The generative
process above is quite flexible and can handle any
directed-acyclic concept graph.  The model cannot, however,
handle cycles in the concept structure as
the walk of the concept graph starting at the root node is not guaranteed to
terminate at an exit node.

The word generation mechanism via the concept route in the hierarchical concept-topic model
is related to the Hierarchical Pachinko Allocation model 2 (HPAM 2) as described in \cite{Mimnohpam07}.  In the HPAM 2 model, topics are
arranged in a 3-level hierarchy with root, super-topics and sub-topics at levels 1,2 and 3
respectively and words are generated by traversing the topic hierarchy and exiting at a
specific level and node.  In our model, we use a similar mechanism but only for word generation
via the concept route.  There is additional machinery in our model to incorporate $T$
data-driven topics (in addition to the hierarchy of concepts) and a
switching mechanism to choose the word generation process via the
concept route or the topic route.

In our experiments, we refer to the hierarchical concept-topic model as HCTM and the version
of the model without topics, which we use for illustrative purposes, as HCM.  Note that
the models we described earlier in Section \ref{sec:ctm} (CM, CTM etc) ignore any
hierarchical information.  There are
several advantages of modeling the concept hierarchy.  We learn the correlations
between the children of a concept via its Dirichlet parameters ($\tau_c$ in
the generative process).  This enables the model to a priori prefer certain paths in the
concept hierarchy given a new document.  For example, when trained on scientific documents
the model can automatically adjust the Dirichlet parameters to give more weight to the child
node ``science" of root than say to node ``society".  We experimentally investigate this
aspect of the model by comparing HCM with CM (more details later).  Secondly, by selecting a
path along the concept hierarchy, the learning algorithm of the hierarchical model also reinforces
the probability of the other concept nodes that lie along the path.  This is desirable since we expect the concepts to
be arranged in the hierarchy by their ``semantic proximity".  We measured
the average minimum path length of
 5 high probability concept nodes for 1000 randomly selected science documents from the TASA corpus for
 both HCM and CM using the CALD concept set. HCM has an average value of 3.92 and CM has an average value of
 4.09, the difference across the 1000 documents is significant under a t-test at the
 0.05 level.  This result indicates that the hierarchical model prefers
 semantically similar concepts to describe documents.  We show some illustrative examples
 in the next section to demonstrate the usefulness of the hierarchical model.

\section{Illustrative Examples } \label{sec:illustrations}

\begin{figure*} \centering
\includegraphics[keepaspectratio,width=5.7in]{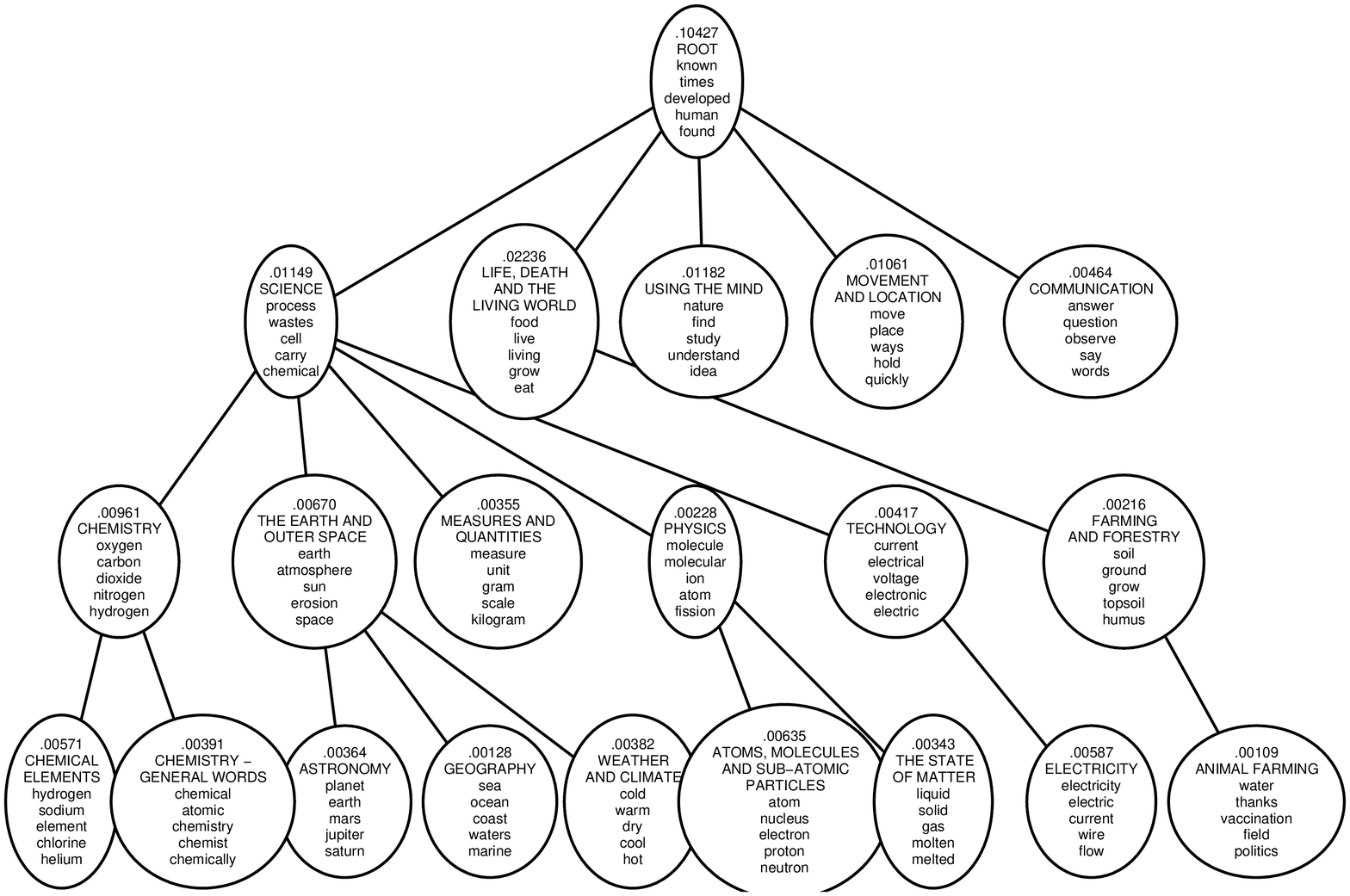}
\caption{Illustrative example of marginal concept distributions from the hierarchical
concept model learned on science documents using CALD concepts. \label{fig:caldmargconcepts}} 
\end{figure*}

\begin{figure*} \centering
\includegraphics[keepaspectratio,width=5.7in]{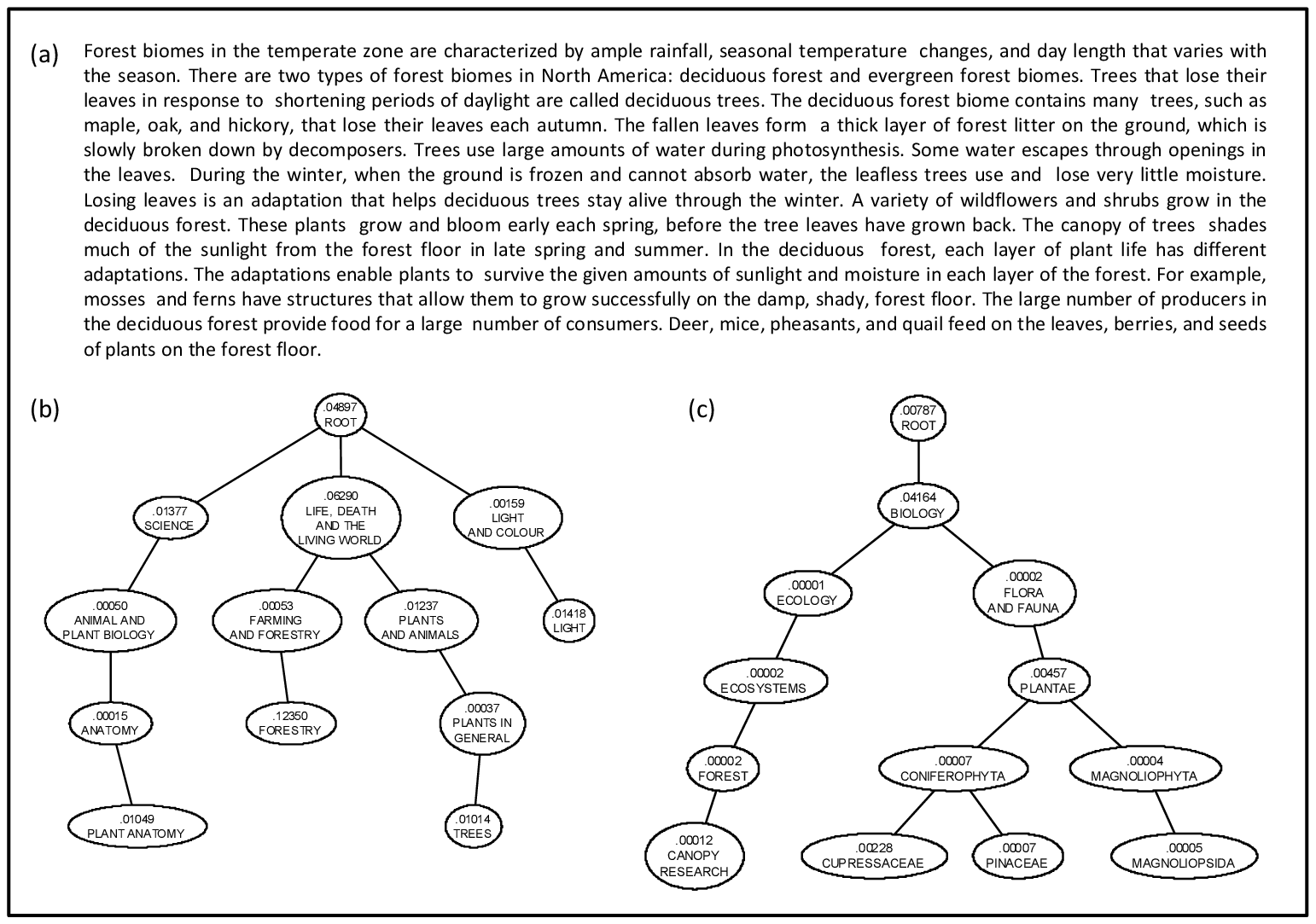}
\caption{Example of a single TASA document from the science
genre (a). The concept distribution inferred by the hierarchical
concept model using the CALD concepts (b) and the ODP concepts (c). \label{fig:taggingexample1}}
\end{figure*}

In this section, we provide two illustrative examples from the
hierarchical concept model trained on the science genre of the
TASA document set. Figure \ref{fig:caldmargconcepts} shows the 20
highest probability concepts (along with the ancestors of those
nodes) for a random subset of 200 documents. The concepts are from the
CALD concept set. For each concept, the name of the concept is
shown in all caps and the number represents the marginal
probability for the concept. The marginal probability is computed
based on the product of probabilities along the path of reaching
the node as well as the probability of exiting at the node and
producing the word, marginalized (averaged) across 200 documents.

Many of the most likely concepts as inferred by the model relate
to specific science concepts (e.g. \textsc{Geography}, \textsc{Astronomy},
\textsc{Chemistry}, etc.). These concepts all fall under the general
\textsc{Science} concept which is also one of the most likely concepts for
this document collection. Therefore, the model is able to
summarize the semantic themes in a set of documents at multiple
levels of granularity. The figure also shows the 5 most likely
words associated with each concept. In the original CALD concept
set, each concept consists of a set of words and no knowledge is
provided about the prominence, frequency or representativeness of
words within the concept. In the hierarchical concept model, for
each concept a distribution over words is inferred that is tuned
to the specific collection of documents. For example, for the
concept \textsc{Astronomy} (second from left, bottom row), the word
``planet" receives much higher probability than the word ``saturn"
or ``equinox" (not shown), all of which are members of the
concept. This highlights the ability of the model to adapt to
variations in word usage across document collections.

Figure \ref{fig:taggingexample1} shows the result of inferring the hierarchical concept mixture for an
individual document using both the CALD and the ODP concept
sets (Figures \ref{fig:taggingexample1}(b) and \ref{fig:taggingexample1}(c) respectively). For the hierarchy visualization, we
selected the 8 concepts with the highest probability and included all
ancestors of these concepts when visualizing the tree. This illustration
shows that the model is able to give interpretable results for an individual
document at multiple levels of granularity. For example, the CALD
subtree (Figure \ref{fig:taggingexample1}(b)) highlights the specific semantic themes of \textsc{Forestry},
\textsc{Light}, and \textsc{Plant Anatomy} along with the more general themes of \textsc{Science} and
\textsc{Life and Death}. For the ODP concept set (Figure \ref{fig:taggingexample1}(c)), the likely concepts
focus specifically on \textsc{Canopy Research}, \textsc{Coniferophyta} and more general
themes such as \textsc{Ecology} and \textsc{Flora and Fauna}. This shows that different
concept sets can each produce interpretable and useful document
summaries focusing on different aspects of the document.

\section{Experiments}  \label{sec:experiments}

We assess the predictive performance of the topic model,
concept-topic model and the hierarchical concept-topic model by
comparing their perplexity on unseen words in test documents using
concepts from CALD and ODP.  Perplexity is a quantitative measure
to compare language models \citep{Brown92} and is widely used to
compare the predictive performance of topic models (e.g.
\cite{BleiNg03,GriffStey04,Chemudugunta07,NCRPBlei07}).  While
perplexity does not necessarily directly measure aspects of a
model such as interpretability or coverage, it is nonetheless a
useful general predictive metric for assessing the quality of a
language model.  In simulated experiments (not described in this
paper) where we swap word pairs randomly across concepts to
gradually introduce noise, we found a positive correlation of the
quality of concepts with perplexity.  In the experiments below, we
randomly split documents from science and social studies genres
into disjoint train and test sets with 90\% of the documents
included in the train set and the remaining 10\% in the test set.
This resulted in training and test sets with $D_{train}$ = 4,820
and $D_{test}$ = 536 documents for the science genre and
$D_{train}$ = 9450 and $D_{test}$ = 1051 documents for the social
studies genre respectively.

\subsection{Perplexity}

Perplexity is equivalent to the inverse of the geometric mean of
per-word likelihood of the heldout data.  It can be interpreted as
being proportional to the distance (cross entropy to be precise)
between the word distribution learned by a model and the word
distribution in an unobserved test document.  Lower perplexity
scores indicate that the model predicted distribution of heldout
data is closer to the true distribution.  More details about the
perplexity computation are provided in the Appendix B.

For each
test document, we use a random 50\% of words of the document to estimate document specific
distributions and measure perplexity on the remaining 50\% of words using the
estimated distributions.

\subsection{Perplexity Comparison across Models}

\begin{figure}
\centering  
\subfigure{\includegraphics[keepaspectratio,width=3.5in]{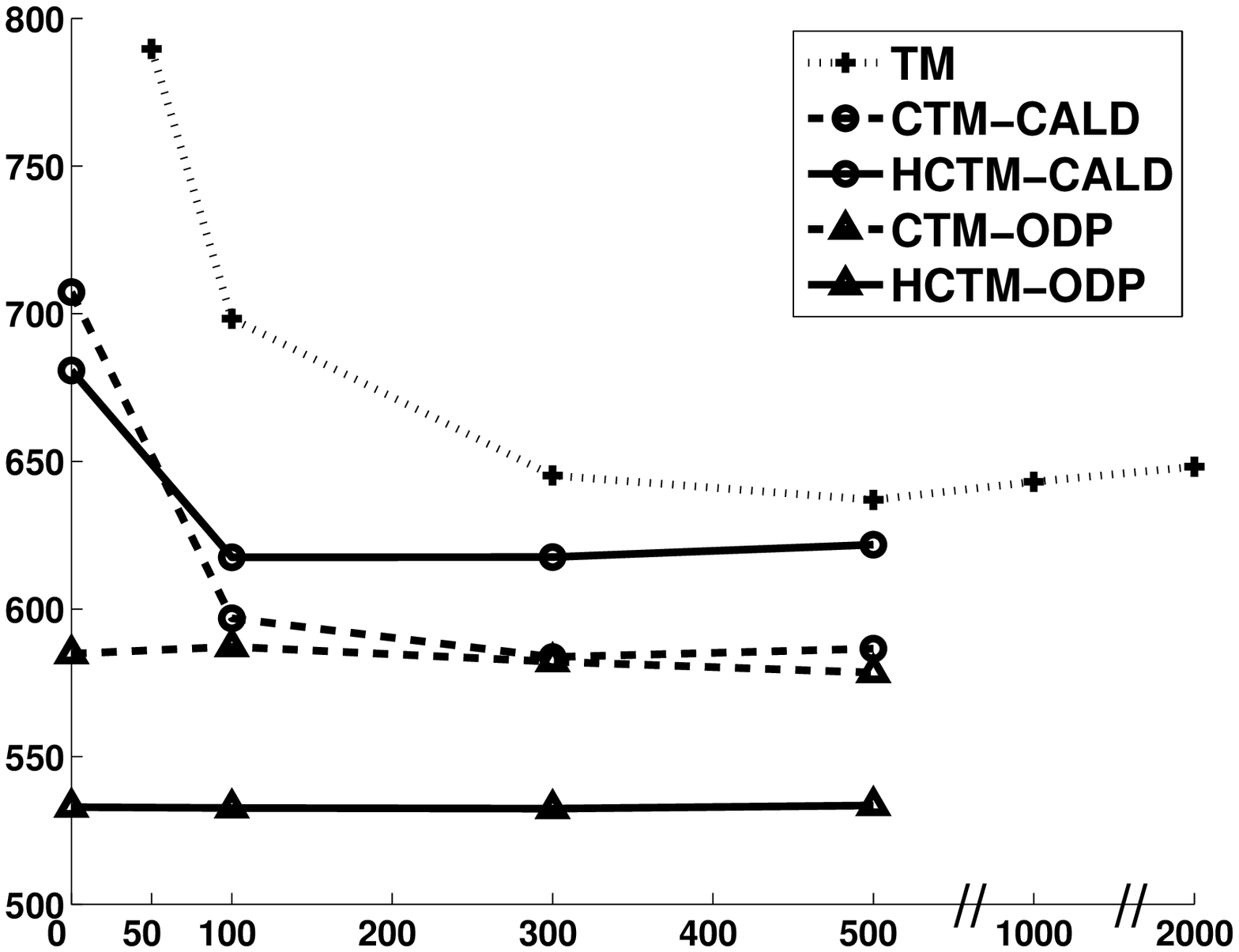}
\label{subfig:combpplextrain5test5}}\hspace{-0.2cm}
\subfigure{\includegraphics[keepaspectratio,width=3.5in]{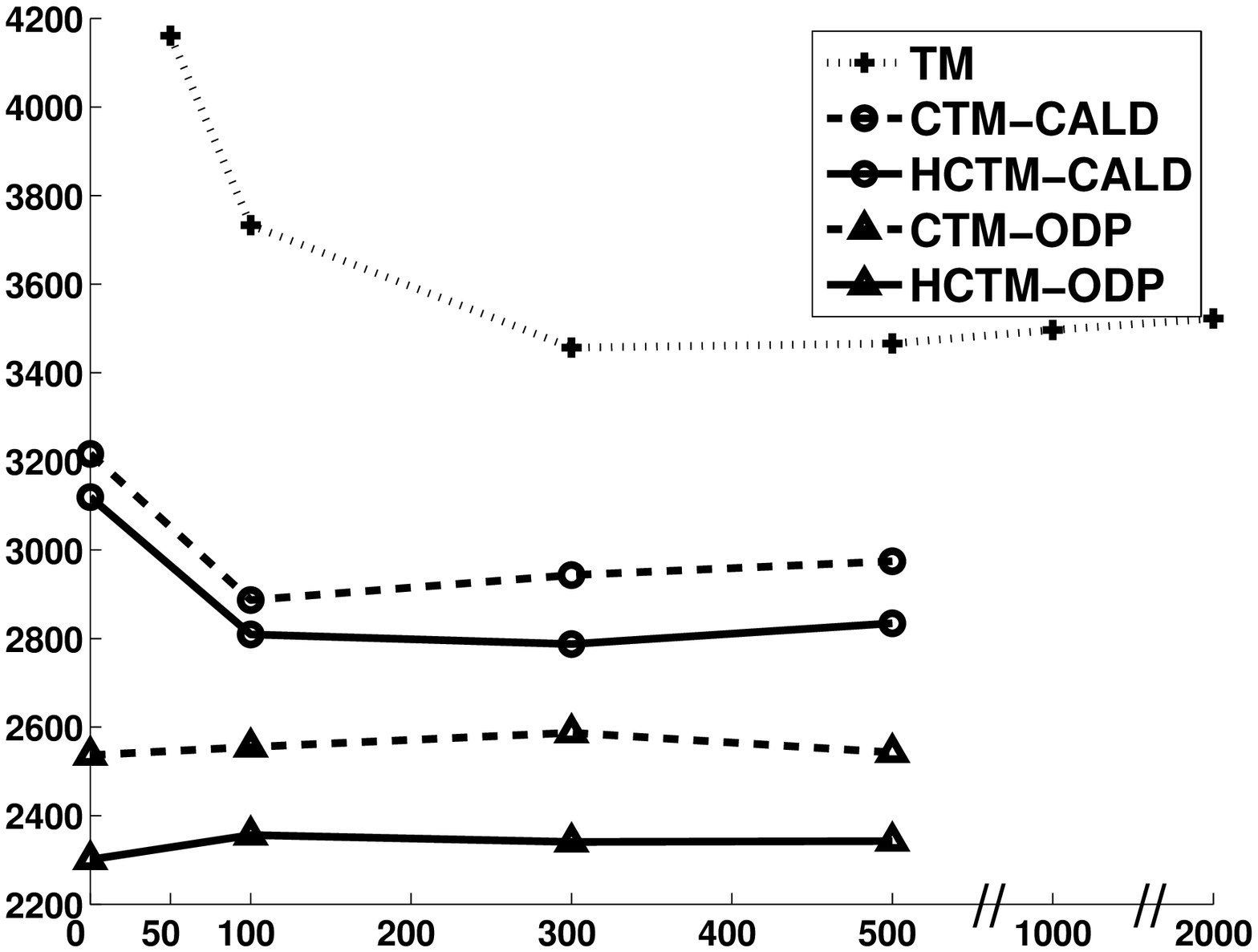}
\label{subfig:combpplextrain5test6}}

\caption{\small\label{fig:combpplextrain5}\small{Comparing perplexity for TM, CTM
and HCTM using training documents from science and testing on science (top) and
social studies (bottom) as a function of number of topics}} 
\end{figure}

\begin{figure}
\centering  
\subfigure{\includegraphics[keepaspectratio,width=3.5in]{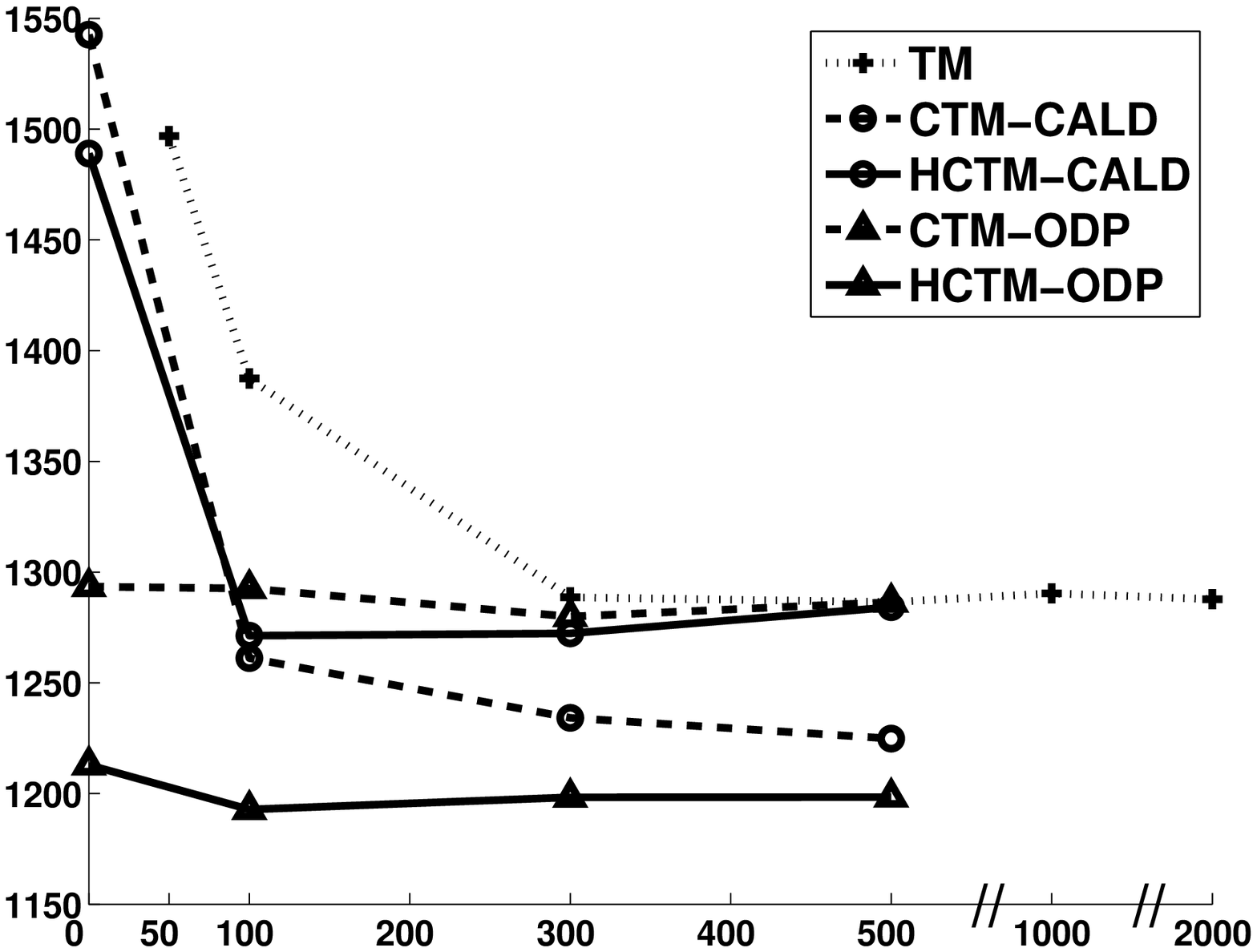}
\label{subfig:combpplextrain6test6}}\hspace{-0.2cm}
\subfigure{\includegraphics[keepaspectratio,width=3.5in]{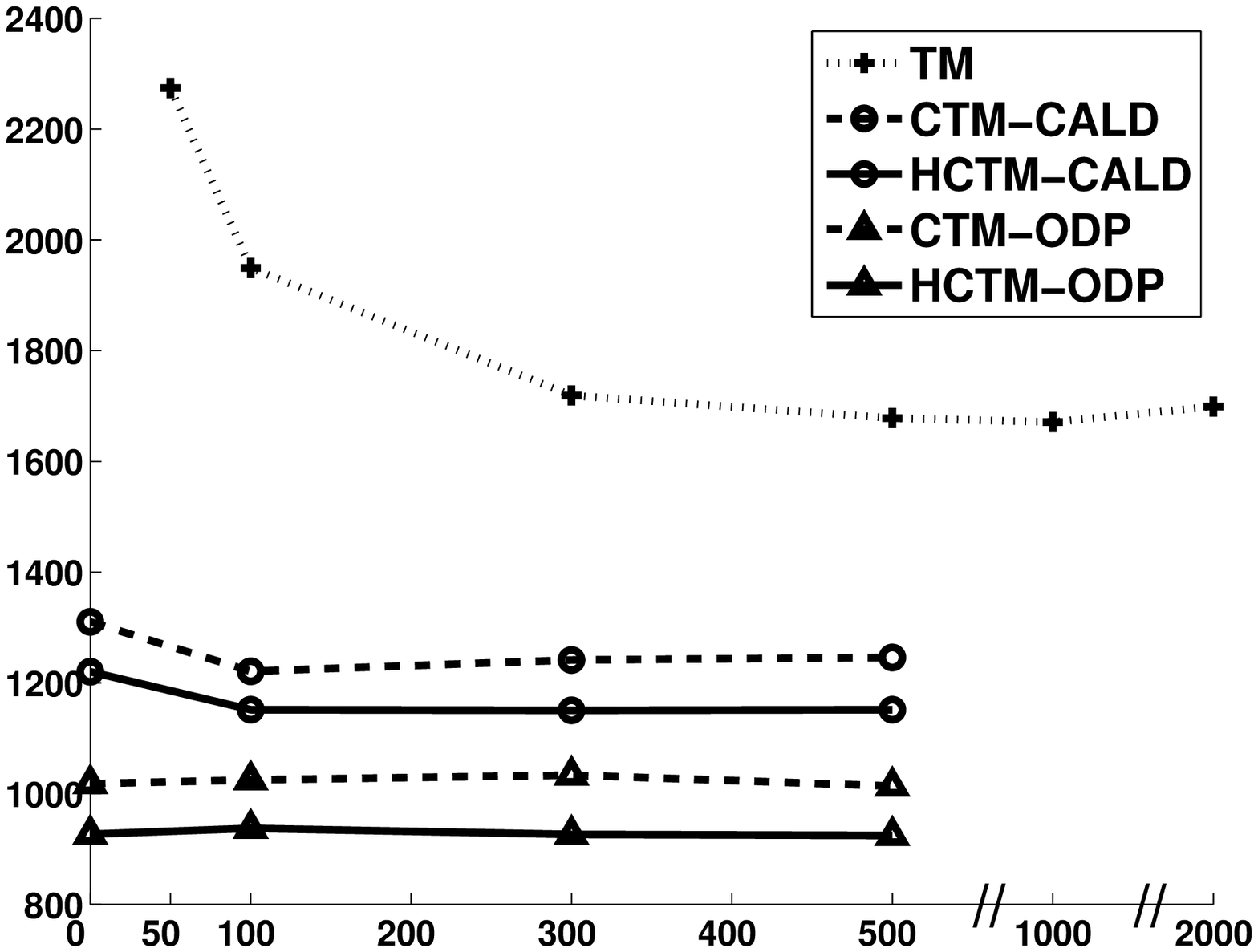}
\label{subfig:combpplextrain6test5}}

\caption{\small\label{fig:combpplextrain6}\small{Comparing perplexity for TM, CTM
and HCTM using training documents from social studies and testing on social studies (top) and
science (bottom) as a function of number of topics}} 
\end{figure}

We compare the perplexity of the topic model (TM), concept-topic model (CTM) and the hierarchical concept-topic
model (HCTM) trained on document sets from the science and social studies genres of the TASA collection and using concepts
 from CALD and ODP concept sets.  For the models using concepts, we indicate the concept set used by appending the
name of the concept set to the model name, e.g. HCTM-CALD to indicate that HCTM was trained using
concepts from the CALD concept set.  Figure \ref{fig:combpplextrain5} shows the perplexity of TM, CTM and HCTM
using training documents from the science genre in TASA and testing on documents from the science (top) and social studies
(bottom) genres in TASA respectively as a function of number of data-driven topics $T$. The point $T$ = 0 indicates
that there are no topics used in the model, e.g. for
HCTM this point refers to HCM.  The results clearly indicate that incorporating concepts greatly improves
the perplexity of the models.  This difference is even more significant when the model is
trained on one genre of documents and tested on documents from a different genre (e.g. see bottom
plot of Figure \ref{fig:combpplextrain5}), indicating that the
models using concepts are robust and can handle noise.  TM, on the other hand, is completely
data-driven and does not use any human knowledge, so it is not as robust.  One
important point to note is that this improved performance by the concept models is not
due to the high number of effective topics ($T+C$).  In fact, even with $T$ = 2,000 topics TM does
not improve its perplexity and even shows signs of deterioration in quality in
some cases.  In contrast, CTM-ODP and HCTM-ODP, using over 10,000 effective topics,
are able to achieve significantly lower perplexity than TM.  The corresponding
plots for models using training documents from
social studies genre in TASA and testing on documents from the social studies (top) and science (bottom) genres in TASA respectively are shown
in Figure \ref{fig:combpplextrain6} with similar qualitative results as
in Figure \ref{fig:combpplextrain5}.  CALD and ODP concept sets mainly contain science-related concepts and do not
contain many social studies related concepts.  This is reflected in the results where the
perplexity values between TM and CTM/HCTM trained on documents from the social studies genre are relatively closer (e.g. as shown in the top plot
of Figure \ref{fig:combpplextrain6}.  This is, of course, not true for the bottom plot as in this case TM again suffers
due to the disparity in themes in train and test documents).

Figures \ref{fig:combpplextrain5} and \ref{fig:combpplextrain6} also allow us to
compare the advantages of modeling the hierarchy of the concept sets.  In both these
figures when $T = 0$, the performance of HCTM is always better than the performance of CTM for
all cases and for both concept sets.  This effect can be attributed to modeling the correlations
of the child concept nodes.  Note that the one-to-one comparison of concept models
with and without the hierarchy to assess the utility of modeling the
hierarchy is not straightforward when $T > 0$ because of the differences in the ways
the models mix with data-driven topics (e.g. CTM could choose to generate 30\% of words
using topics whereas HCTM may choose a different fraction).

\begin{figure}
\centering  
\subfigure{\includegraphics[keepaspectratio,width=3.5in]{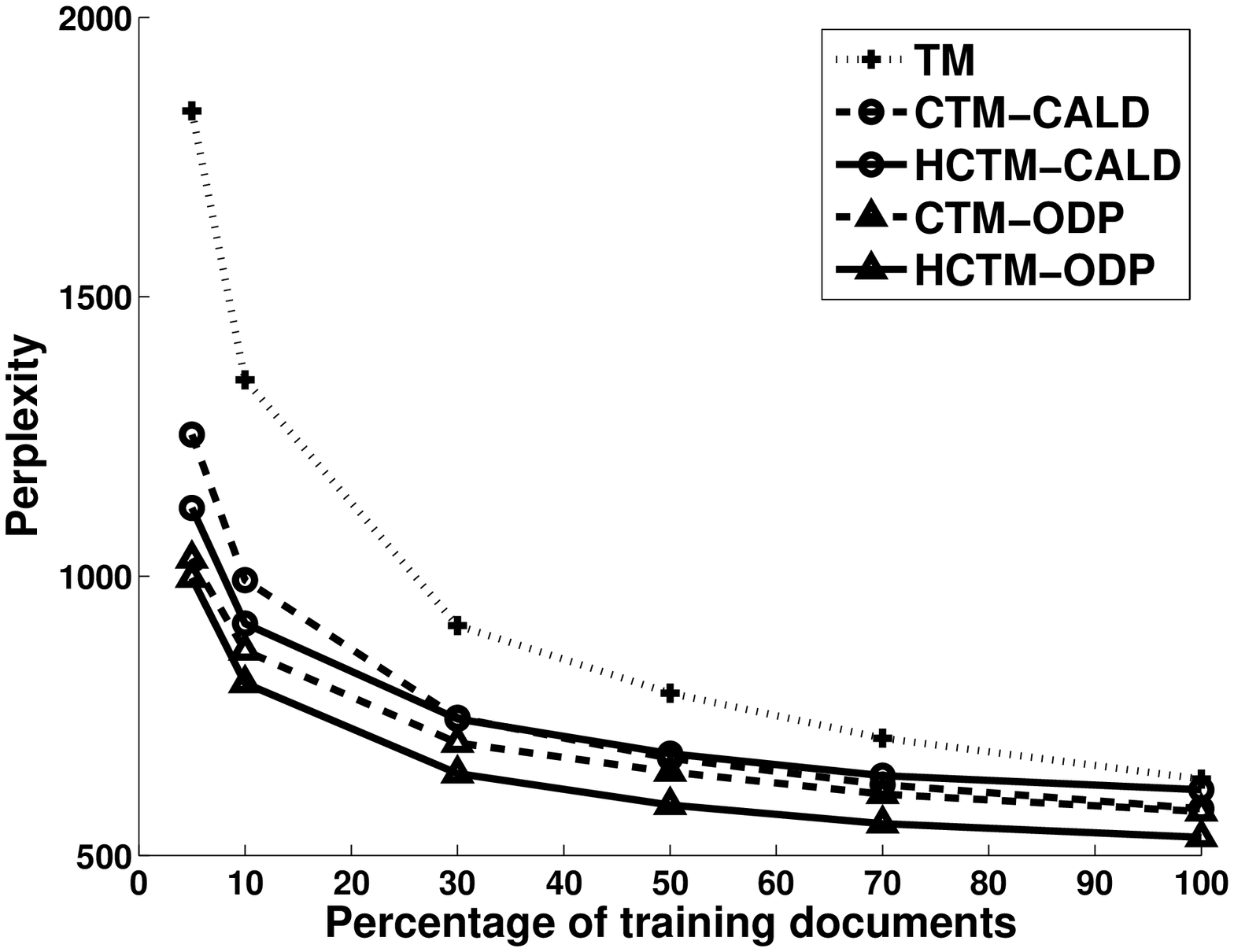}
\label{subfig:percentpplextrain5test5}}\hspace{-0.2cm}
\subfigure{\includegraphics[keepaspectratio,width=3.5in]{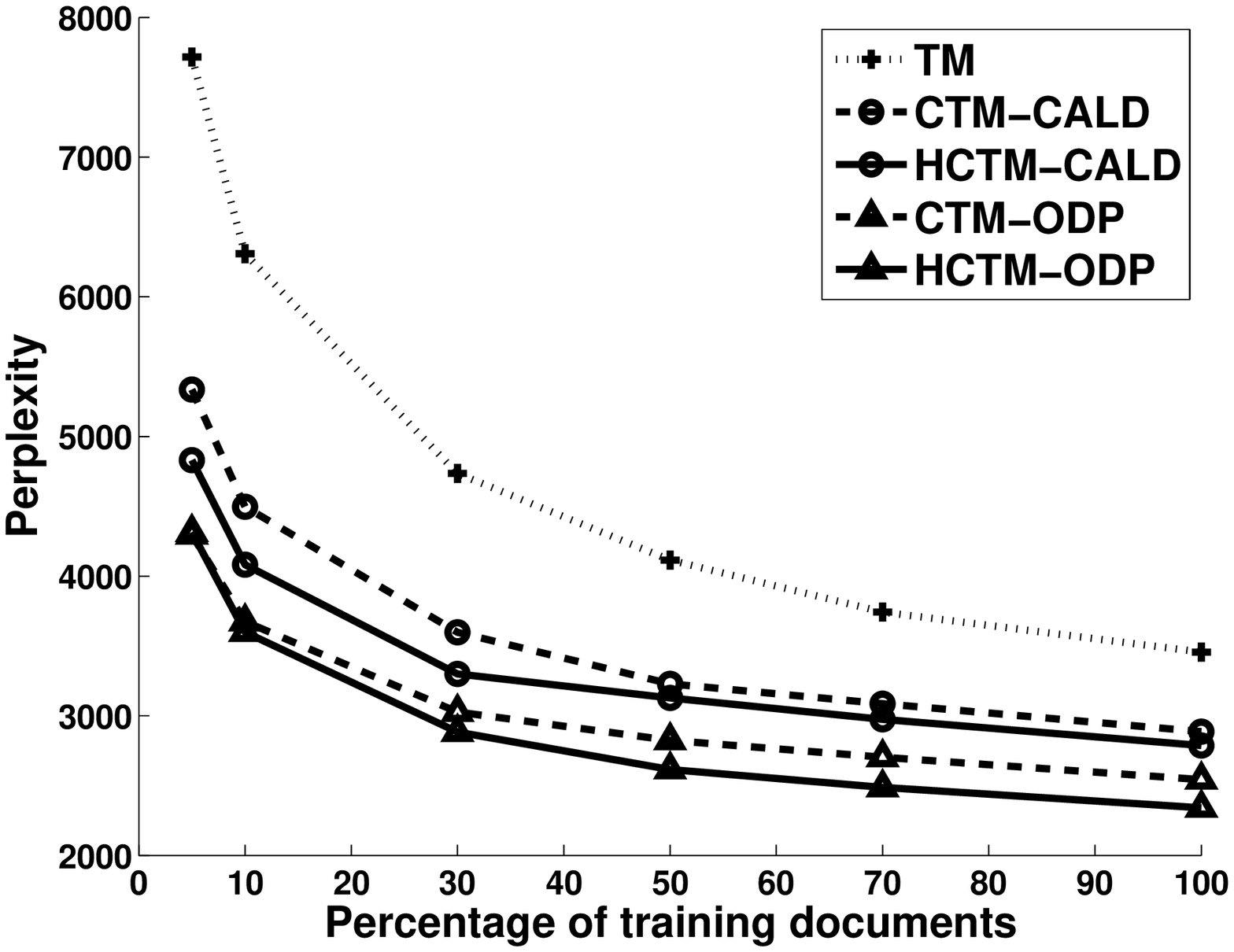}
\label{subfig:percentpplextrain5test6}}

\caption{\small\label{fig:percentpplextrain5}\small{Comparing perplexity for TM, CTM
and HCTM using training documents from science and testing on science (top) and
social studies (bottom) as a function of percentage of training documents}} 
\end{figure}

\begin{figure}
\centering  
\subfigure{\includegraphics[keepaspectratio,width=3.5in]{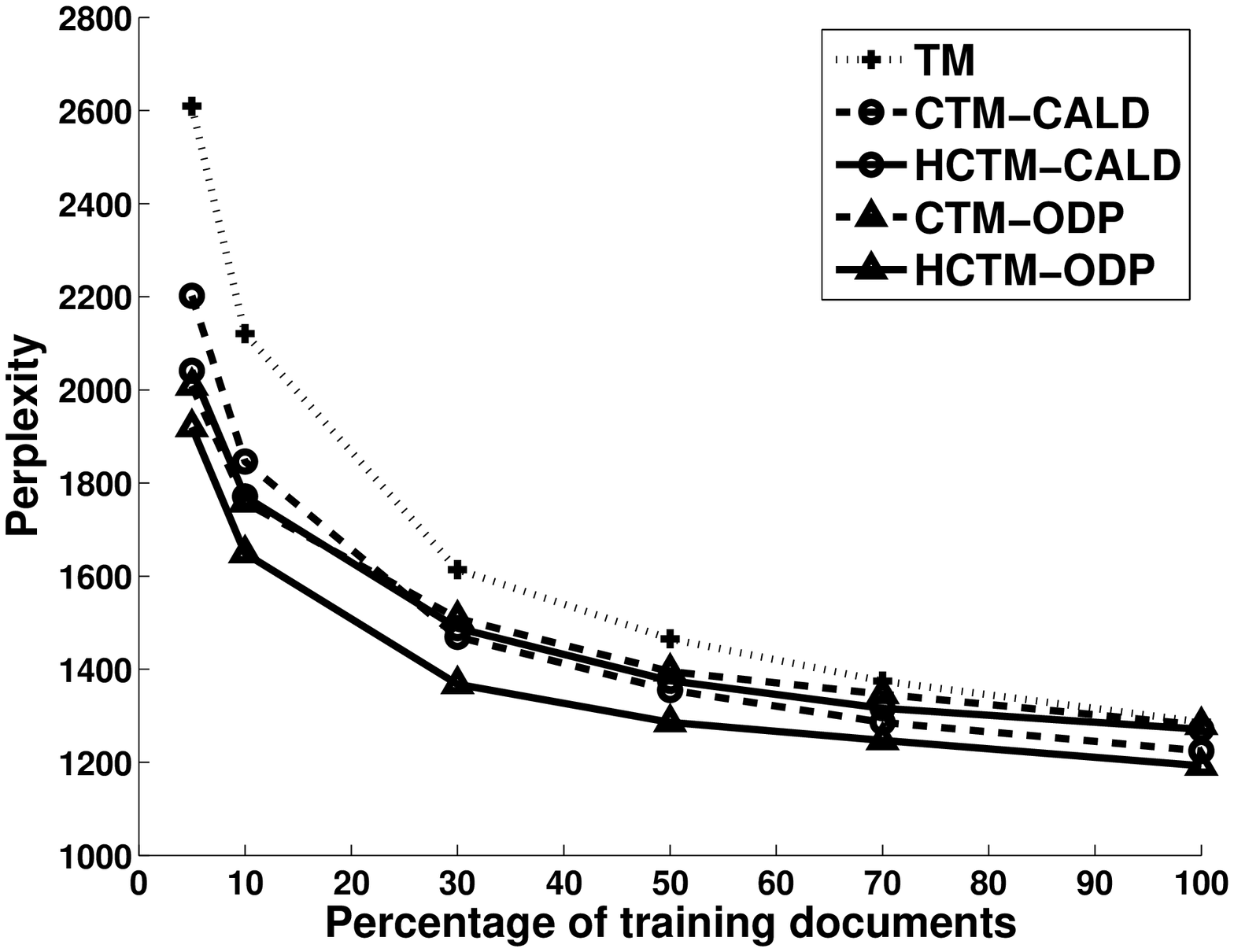}
\label{subfig:percentpplextrain6test6}}\hspace{-0.2cm}
\subfigure{\includegraphics[keepaspectratio,width=3.5in]{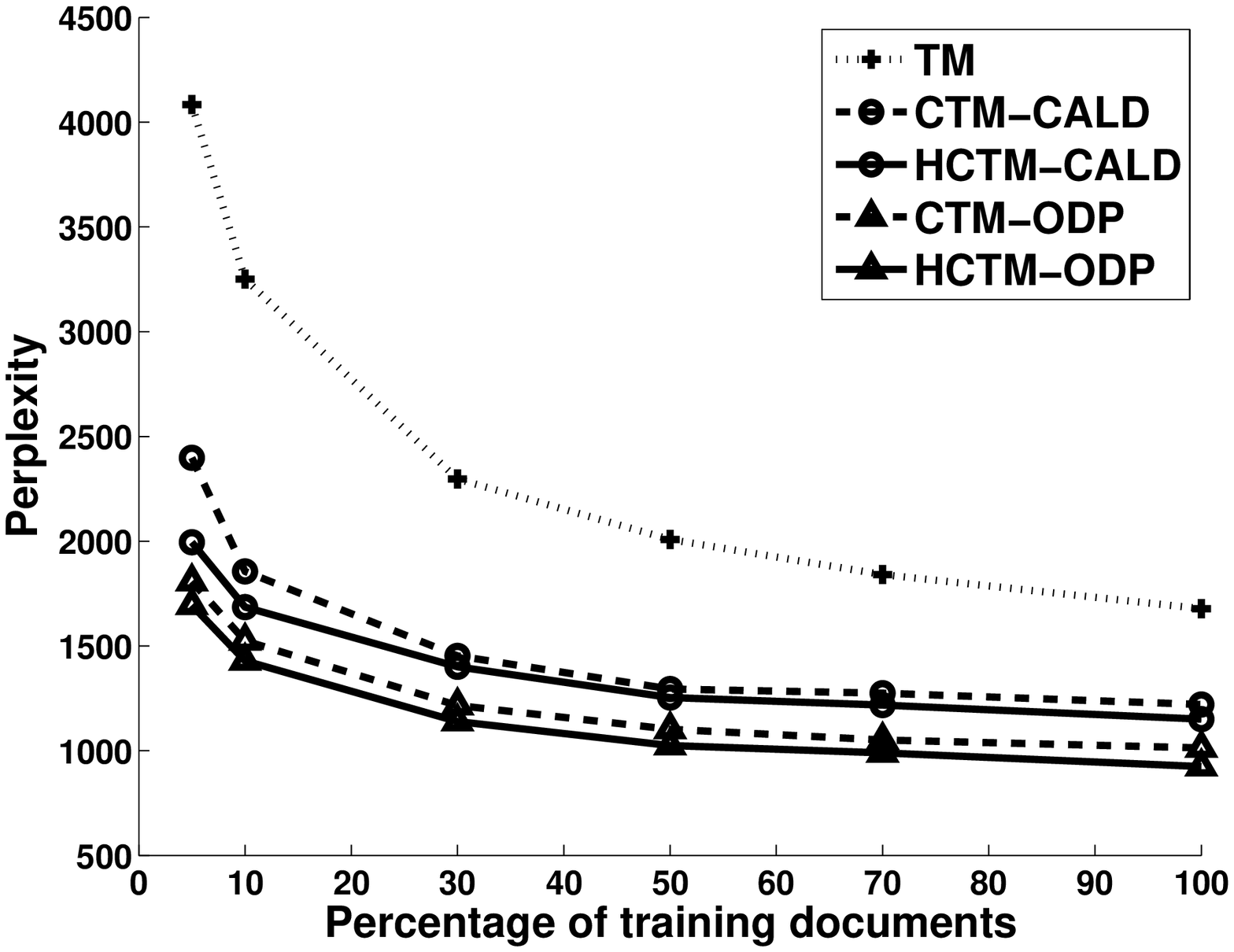}
\label{subfig:percentpplextrain6test5}}

\caption{\small\label{fig:percentpplextrain6}\small{Comparing perplexity for TM, CTM
and HCTM using training documents from social studies and testing on social studies (top) and
science (bottom) as a function of percentage of training documents}} 
\end{figure}

We next look at the effect of varying the amount of training data
for all models.  Figure \ref{fig:percentpplextrain5} shows the
perplexity of the models as a function of varying amount of
training data using documents from the science genre in TASA for
training and testing on documents from the science (top) and
social studies (bottom) genres respectively.  Figure
\ref{fig:percentpplextrain6} shows the corresponding plots for
models using training documents from the social studies genre in
TASA and testing on documents from the social studies (top) and
science (bottom) genres in TASA respectively.  In both these
figures when there is insufficient training data, the models using
concepts significantly outperform the topic model.  Among the
concept models, HCTM  consistently outperforms CTM.  Both the
concept models take advantage of the restricted word associations
used for modeling the concepts that are manually selected on the
basis of the semantic similarily of the words.  That is, CTM and
HCTM make use of prior human knowledge in the form of concepts and
the hierarchical structure of concepts (in the case of HCTM)
whereas TM relies solely on the training data to learn topics.
Prior knowledge is very important when there is insufficient
training data (e.g. in the extreme case where there is no training
data available, topics of TM will just be uniform distributions and will not
perform well for prediction tasks.  Concepts, on the other hand,
can still use their restricted word associations to make 
reasonable predictions). This effect is more pronounced when we
train on on genre of documents and test on a different genre
(bottom plots in both Figures \ref{fig:percentpplextrain5} and
\ref{fig:percentpplextrain6}), i.e. prior knowledge becomes even
more important for this case.  The gap between the concept models
and the topic model narrows as we increase the amount of training
data.  Even at the 100\% training data point CTM and HCTM have
lower perplexity values than TM.

\section{Future Directions } \label{sec:future}

There are several potentially useful directions in which the hierarchical concept-topic
model can be extended.  One interesting extension to try is to substitute the
Dirichlet prior on the concepts with a Dirichlet Process prior.  Under this
variation, each concept will now have a potentially infinite number of children, a
finite number of which are observed at
any given instance (e.g. see \cite{TehJorBea2006}).  When we
do a random walk through the concept hierarchy to generate a
word, we now have an additional option to create a child topic and generate a
word from that topic.  There would be no need for the switching mechanism
as data-driven topics are now part of the concept hierarchy.  This model
would allow us to add new topics to an existing concept set hieararchy and could
potentially be useful in building a recommender system for updating
concept ontologies.

An alternative direction to pursue would be to introduce additional machinery in the generative model to
handle different aspects of transitions through the concept hierarchy.  In HCTM, we
currently learn one set of path correlations for the entire corpus (captured by the Dirichlet parameters $\tau$ in
HCTM).  It would be
interesting to introduce another latent variable to model multiple
path correlations.  Under this extension, documents from different genres can learn
different path correlations (similar to \cite{BoydBlei07}).  For example, scientific documents could prefer to
utilize paths involving scientific concepts and humanities concepts could prefer to
utilize a different set of path correlations when they are modeled together.  This model
would also be able to make use of class labels of documents if available.  Other potential
future directions involve modeling multiple corpora
and multiple concept sets and so forth.

\section{Conclusions}  \label{sec:conclusions}

We have proposed a probabilistic framework for combining data-driven topics and
semantically-rich human-defined concepts.  We first introduced the concept-topic
model, which is a straightforward extension of the topic model, to utilize semantic
concepts in the topic modeling framework.  This model
represents documents as a mixture of topics and concepts thereby allowing us to
describe documents using the semantically rich concepts. We further
extended this model with the hierarchical concept-topic model where
we incorporate the concept-set hierarchy into the
generative model by modeling the parent-child relationship in the
concept hierarchy.

Experimental results, using two document collections and
two concept sets with approximately 2,000 and 10,000 concepts, indicate that using the
semantic concepts significantly improves the quality
of the resulting language models.  This improvement is more pronounced when the training
documents and test documents belong to different genres.  Modeling concepts
and their associated hierarchies appears to be particularly useful when there is
limited training data --- the hierarchical concept-topic model has
the best predictive performance overall in this regime.  We view the current set
of models as a starting point for exploring more expressive
generative models that can potentially have wide-ranging applications, particularly in
areas of document modeling and tagging, ontology modeling and refining, information
retrieval, and so forth.

\section*{Acknowledgements}

The work of Chaitanya Chemudugunta, Padhraic Smyth, and Mark Steyvers was supported in part by the National 
Science Foundation under Award Number IIS-0082489. The work 
of Padhraic Smyth was also supported by a Research Award from Google.

\bibliographystyle{plainnat}
\bibliography{conceptmodel}

\appendix

\section{Inference for the Hierarchical Concept-Topic Model } 
\label{app:hctm}

In this section, we provide more details on inference using collapsed
Gibbs sampling and parameter estimation for the hierarchical
concept-topic model.  For all the models used in the paper (TM, CTM, HCTM etc), we
run Gibbs sampling chains for 500 iterations and estimate the
expected values of the model distributions by averaging over samples from
5 independent chains by collecting one sample from the last iteration of
each chain.  We use a symmetric Dirichlet prior of $0.01$ for the multinomial
distributions over words (i.e. $\beta_{\phi}$ = $\beta_{\psi}$ = $0.01$ where
they are defined) and use asymmetric Dirichlet priors for all the other multinomial
distributions (correspondingly, we use an asymmetric Beta prior $\gamma$ for the Bernoulli
switch distribution $\zeta$ of HCTM) and optimize these parameters using the fixed
point update equations given in \cite{Minka00}.  We update the Dirichlet
parameters after each sweep of the Gibbs sampler through the corpus. \\

In the hierarchical concept-topic model, $\phi$ and $\psi$ correspond to the set of $p(w|t)$
word-topic and $p(w|c)$ concept-topic multinomial distributions with Dirichlet prior $\beta_{\phi}$ and
$\beta_{\psi}$ respectively.  $\xi$ is the set of $p(x|d)$ document-specific
Bernoulli switch distributions with Beta prior $\gamma$.  $\theta$ corresponds to the
set of $p(t|d)$ topic-document multinomial distributions with Dirichlet
prior $\alpha$.  $\zeta_{cd}$ represents the multinomial distribution over the children of
concept node $c$ for document $d$ with Dirichlet prior $\tau_c$ ---for a data set with
$C$ concepts and $D$ documents, there are $C \times D$
such distributions.  Using the collapsed Gibbs sampling procedure, the component
variables $z_i$ and binary switch variable $x_i$ can be efficiently sampled (after marginalizing
the distributions $\phi$, $\psi$, $\xi$, $\theta$ and $\zeta$).  The Gibbs sampling equations
for the hierarchical concept-topic model are as follows: -\\

case (i): ${\bf x}_i = 0$ and $1 \le {\bf z}_i \le T$ \\
\begin{eqnarray}
\small P({\bf x}_i = 0, {\bf z}_i = t | {\bf w}_i = w, {\bf w}_{-i}, {\bf x}_{-i}, {\bf z}_{-i}, \gamma, \alpha, \tau, \beta_{\phi}, \beta_{\psi})
\propto  \\
~~~~({N}_{0d,-i} + \gamma_{0}) \times \frac{ {{C}}_{td,-i} + \alpha_{t} }{ \sum_{t'}{ ( {C}}_{t'd,-i} + \alpha_{t'} ) }  \nonumber \\
~~~~~~~~~~~~~~~~~~~~~~~~~~~~~~~~\times \frac{ {C}_{wt,-i} + \beta_{\phi} }{ \sum_{w'}
{ ( {C}}_{w't,-i} + \beta_{\phi} )}  \label{eqn:samp1} \nonumber
\end{eqnarray} \\
case (ii): ${\bf x}_i = 1$, ${\bf z}_i = T + c$ and $1 \le c \le C$ \\
\begin{eqnarray}
\small P({\bf x}_i = 1, {\bf z}_i = T + c | {\bf w}_i = w, {\bf w}_{-i}, {\bf x}_{-i}, {\bf z}_{-i}, \gamma, \alpha, \tau, \beta_{\phi}, \beta_{\psi})
\propto \\
~~~~({N}_{1d,-i} + \gamma_{1}) \times {\prod_{j=2}^{ |\lambda| } \frac{ {C}_{{\lambda_{j-1}}{\lambda_j}d,-i} + \tau_{{\lambda_{j-1}}{\lambda_j}} }{ \sum_{k}
{ ( {C}}_{{\lambda_{j-1}}kd,-i} + \tau_{{\lambda_{j-1}}{k}} )} } \nonumber \\
~~~~~~~~~~~~~~~~~~~~~~~~~~~~~~~~\times \frac{ {C}_{wc,-i} + \beta_{\psi} }{ \sum_{w' \in c}
{ ( {C}}_{w'c,-i} + \beta_{\psi} )} \nonumber
\end{eqnarray} \\

where $C_{wt}$ and $C_{wc}$ and are the number of times word $w$ is assigned to
topic $t$ and concept $c$ respectively.  $N_{0d}$ and $N_{1d}$ are the number of
times words in document $d$ are generated by topics and by concepts
respectively.  $C_{td}$ is the number of times topic $t$ is associated with
document $d$.  ${\bf \lambda}$ is a vector representing the path from the root
to the sampled concept node $c$ and exiting at $c$ (i.e. $\lambda_1$ is the root, $\lambda_{|\lambda|-1} = c$ and
$\lambda_{|\lambda|}$ is the exit child of concept node $c$).  $ {C}_{{\lambda_{j-1}}{\lambda_j}d}  $ is
the number of times concept node $\lambda_j$ was visited from its parent concept node $\lambda_{j-1}$ in
document $d$.  Subscript $-i$ denotes that the effect of the current word $w_i$ being sampled is removed from the
counts. \\

As mentioned earlier, we use the fixed point update equations described in
\cite{Minka00} to optimize the asymmetric Dirichlet and Beta
distribution parameters.  In the hierarchical concept-topic model,
the Dirichlet distribution parameters $\alpha$ are updates as
follows:

\begin{eqnarray}
\small {\alpha_t}^{new} = {\alpha_t}^{old} \frac{  {\sum_{d} ( \Psi(C_{td} + \alpha_t) - \Psi(\alpha_t) } ) }{
{\sum_{d} ( \Psi(\sum_{t'} C_{t'd} + \sum_{t'} \alpha_t') - \Psi( \sum_{t'} \alpha_{t'}) } ) }
\end{eqnarray} \\

where $\Psi(.)$ denotes the digamma function (logarithmic derivative of the Gamma function). Dirichlet
distribution parameters $\tau_{c}$ for $c \in \{1,...,C\}$ and Beta distribution
parameters $\gamma$ are updated in a similar fashion. \\

Point estimates for the distributions marginalized for Gibbs sampling can be obtained by
using the counts of assignment variables $z_i$ and $x_i$.  The point estimates for $\phi$,
$\psi$, $\xi$, $\theta$ and $\zeta_c$ are given by,

\begin{eqnarray}
E[ \phi_{wt} | {\bf w}, {\bf z}, \beta_{\phi} ] = \frac{ {{C}}_{wt} + \beta_{\phi} }{ \sum_{w'} ( {{C}}_{w't} + \beta{\phi} ) } \label{eqn:estimates}  \\
E[ \phi_{wc} | {\bf w}, {\bf z}, \beta_{\psi} ] = \frac{ {{C}}_{wc} + \beta_{\psi} }{ \sum_{w' \in c} ( {{C}}_{w'c} + \beta{\psi} ) } \nonumber \\
E[ \xi_{xd} | {\bf x}, \gamma ] = \frac{ {{N}}_{xd} + \gamma_x }{ \sum_{x'} ( {{N}}_{x'd} + \gamma_{x'} ) } \nonumber \\
E[ \theta_{td} | {\bf z}, \alpha ] = \frac{ {{C}}_{td} + \alpha_t }{ \sum_{t'} ( {{C}}_{t'd} + \alpha_{t'} ) } \nonumber \\
E[ \zeta_{ckd} | {\bf z}, \tau_c ] = \frac{ {{C}}_{ckd} + \tau_{ck} }{ \sum_{k'} ( {{C}}_{ck'd} + \tau_{ck'} ) } \nonumber
\end{eqnarray} \\ \\

Inference using Gibbs sampling and point estimates for the topic model
and the concept-topic model can be done in a similar fashion. \\

\section{Perplexity} 

Perplexity of a collection of test documents given the training
set is defined as: \\

\[\footnotesize
        \mbox{Perp}({\bf w}_{test}|{\cal D}^{\mbox{train}}) =
        \exp \biggl( -\frac{\sum_{d=1}^{{D}_{test}} \log  p({\bf w}_d | {\cal D}^{\mbox{train}})}
        { \sum_{d=1}^{{D}_{test}} {N_d}}\biggr) \label{eq:perpl}
        \nonumber
 \] \\ \\
 where {${\bf w}_{test}$} is the words in test documents, ${\bf w}_d$ are
 words in document $d$ of the test set, ${\cal D}^{\mbox{train}}$ is the training
 set, and $N_d$ is the number of words in document $d$.

For the hierarchical concept-topic model, we generate sample-based approximations
to $p( {\bf w}_d | {\cal D}^{\mbox{train}})$ as follow: -

\[\small
        p({\bf w}_d | {\cal D}^{\mbox{train}}) ~
        \approx  \frac{1}{S} \sum_{s=1}^S p({\bf w}_d |
        \{\xi^s \theta^s \zeta^s \phi^s\ \psi^s\})
        \nonumber
 \] \\ 

where $\xi^s$, $\theta^s$, $\zeta^s$, $\phi^s$ and $\psi^s$ are point estimates from $s$ = $1:S$
different Gibbs sampling runs as defined in Appendix \ref{app:hctm}.  Given these point
estimates from $S$ chains for document $d$, the probability of the words ${\bf w}_d$ in document
$d$ can be computed as follows:

\[ p({\bf w}_d |   \{\xi^s \theta^s \zeta^s \phi^s\ \psi^s\ \}) \ = ~~~~~~~~~~~~~~~~~~~~~~~~ \]

\[ ~~~~~~~~~~~~~~~~   \prod_{i=1}^{N_d} \left[
       \xi^s_{0d} \sum_{t=1}^T   \phi^s_{w_it} \theta_{td}^s + \\ \xi^s_{1d} \sum_{c=1}^C \psi_{w_ic}^s
       \prod_{j=2}^{|\lambda|} \zeta^s_{{\lambda_{j-1}}{\lambda_j}d}
        \right]
  \] \\

 where $N_d$ is the number of words in the test document $d$ and $w_i$ is the $i$th word being
 predicted in the test document and $\lambda$ represents a path to the exit 
 child of concept node $c$, starting at the root 
 concept node.  Perplexity can be computed for the topic model and
 the concept-topic model by following a similar procedure.

\end{document}